\documentclass{article}

\usepackage{PRIMEarxiv}

\usepackage[utf8]{inputenc} % allow utf-8 input
\usepackage[T1]{fontenc}    % use 8-bit T1 fonts
\usepackage[hyphens]{url}
\usepackage{hyperref}        % hyperlinks          % simple URL typesetting
\usepackage{booktabs}       % professional-quality tables
\usepackage{amsfonts}       % blackboard math symbols
\usepackage{nicefrac}       % compact symbols for 1/2, etc.
\usepackage{microtype}      % microtypography
\usepackage{lipsum}
\usepackage{fancyhdr}       % header
\usepackage{graphicx}       % graphics
\graphicspath{{media/}}     % organize your images and other figures under media/ folder

\usepackage{amssymb}
\usepackage{graphicx}
\usepackage[acronym]{glossaries}
\usepackage{glossary-inline}
\usepackage{amsmath}
\usepackage{subcaption}
\usepackage{lscape} 
\usepackage{booktabs}
\usepackage{multirow}
\usepackage[official]{eurosym}

\title{A Reinforcement Learning Approach for the Continuous Electricity Market of Germany: Trading from the Perspective of a Wind Park Operator}

\author{
  Malte Lehna, Björn Hoppmann, René Heinrich,Christoph Scholz \\
  Fraunhofer Institute for Energy Economics and Energy System Technology (IEE),\\
  K{\"o}nigstor 59, 34119 Kassel, Germany\\
  \texttt{\{malte.lehna,rene.heinrich,christoph.scholz\}@iee.fraunhofer.de,bjoernhoppmann@web.de} \\
  }

\makeglossaries
\newacronym{rl}{RL}{Reinforcement Learning}
\newacronym{drl}{DRL}{Deep Reinforcement Learning}
\newacronym{ppo}{PPO}{Proximal Policy Optimization }
\newacronym{id}{ID}{Intraday}
\newacronym{da}{DA}{Day-Ahead}
\newacronym{re}{RE}{Renweable Energie}
\newacronym{ml}{ML}{Machine Learning}
\newacronym{pbt}{PBT}{Population Based Training}
\newacronym{lob}{LOB}{Limit Order Book}
\newacronym{ptd}{p.t.d}{prior to delivery}
\newacronym{mdp}{MDP}{Markov Decision Process}
\newacronym{rnn}{RNN}{Recurrent Neural Network}
\newacronym{sgd}{SGD}{Stochastic gradient decent}
\newacronym{a3c}{A3C}{Asynchronous Advantage Actor-Critic}
\newacronym{lstm}{LSTM}{Long Short-Term Memory}

\begin{document}
\maketitle

% keywords can be removed

\begin{abstract}
%~275 Words
With the rising extension of renewable energies, the intraday electricity markets have recorded a growing popularity amongst traders as well as electric utilities to cope with the induced volatility of the energy supply. Through their short trading horizon and continuous nature, the intraday markets offer the ability to adjust trading decisions from the day-ahead market or reduce trading risk in a short-term notice. Producers of renewable energies utilize the intraday market to lower their forecast risk, by modifying their provided capacities based on current forecasts. 
However, the market dynamics are complex due to the fact that the power grids have to remain stable and 
electricity is only partly storable. Consequently, robust and intelligent trading strategies are required that are capable to operate in the intraday market. In this work, we propose a novel autonomous trading approach based on \gls{drl} algorithms as a possible solution. For this purpose, we model the intraday trade as a \glsfirst{mdp} and employ the \glsfirst{ppo} algorithm as our \gls{drl} approach. A simulation framework is introduced that enables the trading of the continuous intraday price in a resolution of one minute steps. We test our framework in a case study from the perspective of a wind park operator. We include next to general trade information both price and wind forecasts. On a test scenario of German intraday trading results from 2018, we are able to outperform multiple baselines with at least 45.24\% improvement, showing the advantage of the \gls{drl} algorithm. However, we also discuss limitations and enhancements of the \gls{drl} agent, in order to increase the performance in future works. 
\end{abstract}

\section{Introduction}
\label{sec:introduction}
\printglossary[type=\acronymtype,style=inline]
%\printglossary
\subsection{Overview}
\label{ssec:motivation}
In past decades, the day-ahead electricity markets were the major trading location to auction electricity capacities between market participants. However, with the European Green Deal \cite{greendeal}
and the rising proportion of \gls{re} production \cite{renewableenergies}, there is a need for a higher flexibility in the electricity trade. Both electricity providers and prosumers require short-term solutions to respond to market changes and reduce their overall trading risk \cite{kiesel2017econometric,shinde2019literature}. One possible solution is the \gls{id} market that enable trading with less than one day prior to the delivery. Here, especially \gls{re} producers can anticipate changes in their production forecast and modify their orders from the day-ahead market to ensure an accurate delivery. 
Consequently, the \gls{id} electricity markets have an increasing popularity amongst traders, resulting in record trading volumes across all European \gls{id} electricity spot markets in 2020\cite{epextrade}. Similarly, in the research community there has been a rising interest in the analysis of the \gls{id} trade. The \gls{id} trade was examined from various perspectives, given that both the expansion of \gls{re} as well as the design of the \gls{id} markets differ across nations.%\footnote{Some markets were favoured by researchers (Spanish, German and Nordic \gls{id} market), yet other markets (Australian and UK \gls{id} market) were analyzed as well \cite{shinde2019literature}.}
In many cases the \gls{id} markets have been analyzed from a structural point of view, e.g., for the identification of influencing factors \cite{ziel2017modeling}. However, actual trading strategies have not been part of many analysis, considering that \gls{id} trading is not comparable to other commodity markets. With its short-term perspective, the strong influence of the \gls{re}, the volatility and the stochastic nature of the \gls{id} market, traditional trading methods are often not applicable. Consequently, alternative methods have to be investigated that can cope with these difficulties.
\\ In our work, we want to contribute to this open question, by proposing an autonomous trading agent based on \glsfirst{drl}. While the autonomous approaches is primarily designed to compensate forecast errors on the \gls{id} market, it also contribute to more system stability. Especially in combination with  flexible energy providers, autonomous agents could be deployed in the long run to counteract against potential surpluses and thus reduce the risk of shortfalls in the electricity distribution. In order to ensure a realistic setting, the autonomous agent is deployed from the perspective of an electricity provider, specifically a wind park operator. Given that wind forecasts always contain some uncertainty, wind park operators have to incorporate this uncertainty in their production plan. Hence, they can not exactly determine their produced volume a priori. The autonomous agent has to adjust the traded volumes on the \gls{id} markets based on recent changes in the forecasts.

\subsection{Related Work}
\label{ssec:rel_work}
In terms of the research community, various topics regarding the \gls{id} market have been publicized. Many of them can be categorized either in the field of structural analysis, price forecasting or in the field of \gls{id} trading. In the first category, papers have been published that examined and analyzed different influencing factors on the \gls{id} markets \cite{kiesel2017econometric,koch2019short,ziel2017modeling,kath2019modeling,pape2016fundamentals}. Naturally, most of the papers reviewed the variability of the \gls{id} trade and were sometimes combined with an analysis of the \gls{re} or their forecasting errors \cite{ziel2017modeling}. However, these analysis focused in most cases on a longer time horizon, thus are not always suited for a direct application. In contrast, the second topic of electricity price forecasting considers the short-term relationship between prices and their influencing factors, as seen in \cite{narajewski2019econometric,narajewski2020ensemble,uniejewski2019,janke2019forecasting,scholz2020towards}. However, next to differences in the markets and the forecast models, the research often differed in the frequency of the forecast target. As example, \cite{narajewski2019econometric} predicted the ID3 index of the German \gls{id} market, while \cite{narajewski2020ensemble,scholz2020towards} predicted the price movement of individual products. As a consequence, different influencing factors were of importance for the different researchers.\\
Lastly, the third research subject is the trading of electricity and its automation on the \gls{id} market, where for example \gls{ml} approaches were frequently published by different researchers, e.g., \cite{wang2019machine,baltaoglu2018algorithmic}. In recent years, a new \gls{ml} approach was proposed that is especially suited to cope with the volatile price progression and the stochastic nature of the \gls{id} market. This respective approach is \gls{drl}, which can model complex relations and at the same time does not necessarily require expert knowledge of the trade relations. Accordingly, there have been first publications for both the \gls{da} and \gls{id} market that addressed the challenges of the electricity trade in combination with \gls{drl}. In terms of the \gls{da}, the researchers utilized the \gls{drl} benefits to automatize and optimize the bidding process, sometimes in combination with the \gls{id} market. These bidding processes could be from the perspective of one \cite{ye2019deep} or multiple \cite{rashedi2016markov} producers, a load serving entity \cite{xu2019deep} or a electric storage unit \cite{wang2018energy}. For the \gls{id} market, especially the papers of \cite{bertrand2018analysis,boukas2018intra,bertrand2019reinforcement} are noteworthy, which proposed trading strategies for the German continuous \gls{id} market. All three papers based their strategy on a \gls{mdp}, however differed in their algorithmic implementation and their research objective. In the paper of  \cite{bertrand2018analysis}, \gls{rl} was used to implement a threshold policy for accepting price bids in the \gls{id} market. To determine the optimal policy, they defined an one-stage \gls{mdp} and used the REINFORCE algorithm to solve the problem. A different use-case was proposed by \cite{boukas2018intra,bertrand2019reinforcement}, who both analyzed the trading under the consideration that they need to optimize the usage of an energy or pump storage unit. Through the assumption that the electricity can be stored, they were able to formulate the trading in a multi-stage \gls{mdp}. In terms of \cite{boukas2018intra}, a Deep-Q network was implemented as \gls{drl} method, while  \cite{bertrand2019reinforcement} again applied the REINFORCE algorithm. 
Even though \cite{bertrand2019reinforcement} showed the most advanced modelling of the \gls{id} trade, they still needed to impose some simplifications to reduce the dimensionality of the optimization problem.
Accordingly, they incorporated price thresholds to limit the number of available orders and traded in 24 time steps for each day. By reducing the number of observations per episode, \cite{bertrand2019reinforcement} were able to implement their \gls{drl} approach with satisfying results. However, under consideration that the \gls{id} trade is indeed traded in milliseconds, there is still a need to improve the \gls{drl} to make it applicable for the \gls{id} trade. In our work, we  address this problem and offer the next step in the development of an autonomous trading agent. 

\subsection{Research Contribution}
\label{ssec:res_contr}
As outlined in the related work section, there are still unanswered questions concerning the \gls{id} electricity trade. Especially, the continuous nature and the resulting shortest-term perspective of the electricity trade has not been part of an investigation. However, given the increasing number of trades and the induced uncertainty of the \gls{re}, there is a practical need to analyze the capabilities of autonomous trading. Therefore, the contributions of this paper can be outlined as follows: 

\begin{enumerate}
    \item  We model the \gls{id} market as a (first-order) \gls{mdp} and propose \gls{drl} approach to simulate the continuous \gls{id} trade.
    \item We develop a \gls{drl} environment based on the OpenAI library \cite{OpenAI} with a shortest-term perspective in one minute resolution. To the best of our knowledge, this  has not been part of any analysis and offers new insights to the electricity trade. 
    \item We validate our proposed \gls{drl} problem from the perspective of an electricity producer that trades the capacities of his wind turbine on the German \gls{id} market.
    \item Within our case study, we are able to show that the proposed \gls{drl} agent is able to trade volatile wind volumes in the \gls{id} simulation.
    \item The \gls{drl} agent outperforms the best baseline agent with 45.24\% more net profit in our case study.  
\end{enumerate}
With the outlined research contribution, the remaining paper is structured as follows. In Section \ref{sec:method}, the methodology of the underlying \gls{drl} is presented. In Section \ref{sec:framework} we transfer the framework to the scenario of the \gls{id} spot market. Thereafter, we  elaborate our case study in Section \ref{sec:experiment} and present and discuss the results in Section \ref{sec:results}. Lastly, in Section \ref{sec:conclusion} we give an overall conclusion. 

%\newpage
\section{Methodology}
\label{sec:method}
\subsection{Reinforcement Learning as Markov Decision Process}
\label{ssec:mdp}
\begin{figure*}[t]
    \centering
    \includegraphics[trim={0cm 11cm 1cm 4.5cm}, clip, scale=0.80]{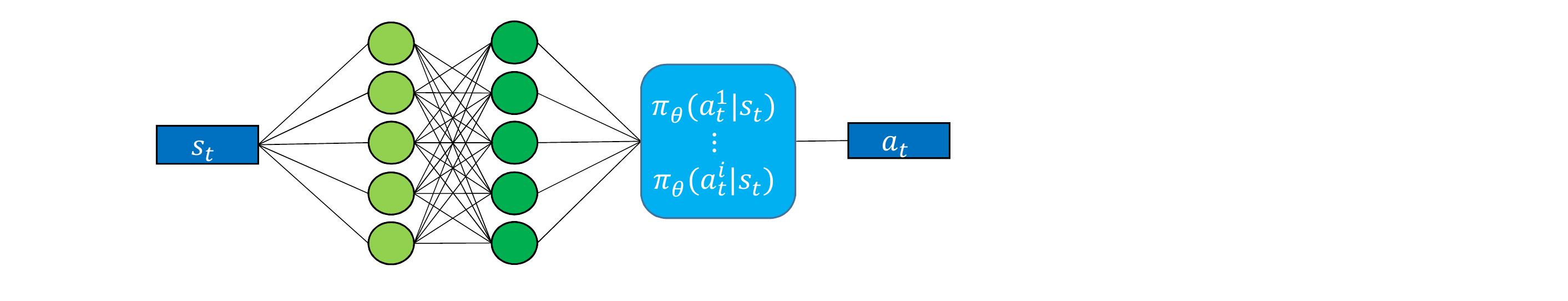}
    \caption{Exemplary neural network of the \gls{drl} approach. The hidden layers (green) pass the state $s_t$ and reward $r_t$ information to the policy layer (light blue). Thereafter, an action $a_t$ is chosen based on the policy.}
    \label{fig:nn}
\end{figure*}

From the area of machine learning methods, the \glsfirst{rl} is an approach first introduced by \cite{sutton1999policy}. It is based on the concept that a model can solve complex problems without outlining the exact relations but instead by learning the correct behavior through incentives in the training process.
\begin{figure*}[ht]
    \centering
    \includegraphics[trim={0cm 9cm 1cm 3.5cm}, clip, scale=0.5]{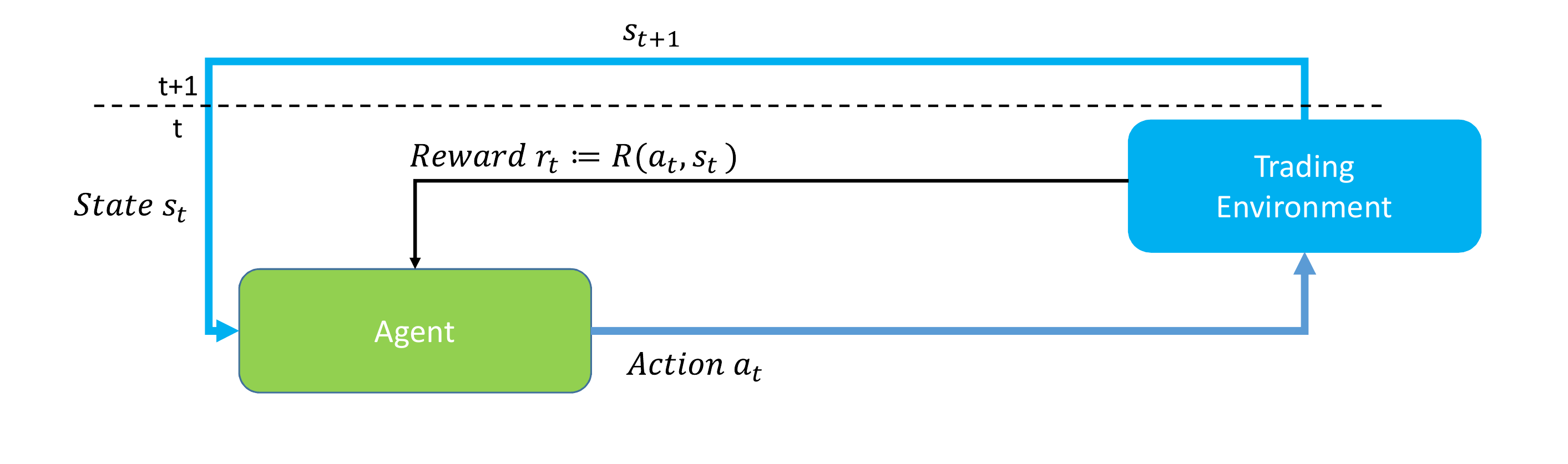}
    \caption{Visualization of the \gls{drl}-Framework with environment and agent. Based on the state $s_t$  of the environment, the agent computes the action $a_t$. The action is returned, the reward $r_{t}$ calculated and saved for the optimization. Thereafter, the environment transitions to the new state $s_{t+1}$.}
    \label{fig:overview}
\end{figure*}
In the \gls{rl} approach, an agent is trained in an environment $\mathbb{S}$ that incorporates the structure of the real-life problem. The environment iterates over consecutive steps, as seen in Figure \ref{fig:overview}, until an episode is complete.\footnote{Depending on the design of the environment, the length of an episode is either fixed or is subject to the actions of the agent.} For each step of the episode, the agent can take different actions $a \in \mathbb{A}$, with $\mathbb{A}$ denoting the space of possible actions.\footnote{Most common are either a discrete action space, with $a$ consisting of specific actions, e.g., in \cite{mnih2015human}, or continuous action space, where $a$ consist of a continuous interval, e.g., in \cite{lillicrap2015continuous}. } 
The purpose of the environment is the communication with the \gls{rl} agent, by supplying the information on its current state $s_t \in \mathbb{S}$ at time $t$ and returning a feedback on the actions $a_t$ in form of a reward $r \in \ \mathbb{R}$ with $r_{t} := R(a_t, s_t)$, based on the reward function $R: \mathbb{A} \times \mathbb{S} \to \mathbb{R}$.
Within the episode of length $T$, the target of the \gls{rl} algorithm is to maximize the cumulative reward $\tilde{R}_{cum} := \sum_{t=1}^T r_t$. In each step, the agent consequently chooses the action, which promises the best possible payout, considering all future actions. 
To correctly define the impact of each action, all the future reward arising from the action has to be considered.  For this purpose, the \gls{rl} problem is assumed to be a \gls{mdp}, i.e., a (time-discrete) stochastic process $\{ M \}_{t \in T}$ on a probability space $(\Omega, \mathcal{A}, \mathbb{P})$ with $t$ denoting the set of time steps $\{ 0, 1, \ldots,T\}$, fulfilling the first-order Markov property for all $t \in T$, and all state-action pairs $(a_t, s_t) \in \mathbb{A} \times \mathbb{S}$:
\small{
\begin{equation}
    \mathbb{P}(s_{t+1}|a_t, s_t) = \mathbb{P}(s_{t+1}|a_t, s_t, \ldots, a_0,s_0 ).
\end{equation}
}
Here, $\mathbb{P}$ is the transition probability to reach state $s_{t+1}$ from state $s_t$ taking action $a_t$.
In a first-order \gls{mdp}, not all past information about the process are required to asses the probability, but instead the current state $s_t$ and the action $a_t$ are sufficient enough.
The consequence of assuming a \gls{mdp} for the \gls{rl} approach is that the \gls{rl} algorithms need to approximate the underlying transition probability $\mathbb{P}$ and selects actions, which result in states leading to a higher reward.
In \gls{rl} terms, policies $\pi_{\theta}$ are probability distributions in each state $s_t$ over all the possible actions $a_t \in \mathbb{A}$. Each probability $\pi(\cdot|s_t)$ denotes the likelihood of the respective action to yield the maximum reward over the whole episode. Note that next to the current state $s_t$, the policies depend on an underlying parameter $\theta$. In the \glsfirst{drl} approach, first introduced by \cite{mnih2015human}, these parameters are the weights of a deep neural network, representing the policy and thereby the action selection behavior. Through optimizing the weights of the network, \gls{drl} algorithms aim to approximate the optimal policy $\pi^* := \pi_{\theta^*}$ offering the best decision for each state $s_t \in \mathbb{S}$. With regard to each step, displayed in Figure \ref{fig:nn}, the network receives the current state $s_t$, as well as the previous reward $r_t$ for the policy optimization. The state information is forwarded to compute the policies $\pi_{\theta}(a^i_t|s_t)$ of each available action $a^i_t$. Based on the best policy, an action $a^*_t$ is chosen, the reward $r_{t+1} = R(a^*_t, s_t)$ is computed and the next iteration step is started. 

\subsection{Optimization Algorithm}
\label{ssec:algorithm}
There are various \gls{drl} algorithms available to update the policy $\pi(\theta)$ that mainly differ in their update strategy and the underlying relationship between policy $\pi(\theta)$ and parameters $\theta$. In our work, we chose the \gls{ppo} algorithm because it enables the usage of an continuous action space. It was firstly introduced in \cite{schulman2017proximal} and is an enhancement of the \gls{a3c} method \cite{mnih2016asynchronous}.
In the \gls{ppo} method, two neural networks are utilized to optimize the policy iteratively. The \textit{Actor} model is deployed in the environment to generate observations by computing actions and obtaining  rewards. The second model, i.e., the \textit{Critic}, forecasts based on the observations the probable future rewards given a specific action. The difference between the forecasted and actual reward is then evaluated in an advantage function  $\hat{A}_t(\theta)$. Actions leading to a positive advantage (so higher reward gained by \textit{Actor} than forecasted by \textit{Critic}) are reinforced while actions leading to negative advantage are discarded. This optimization problem is formulated as a loss function $L(\theta)$, with the objective to minimize the loss based on the expected return and the parameters $\theta$ of the \gls{drl} model:
\begin{align}
    L_t(\theta)&=\hat{E}_t\left[\log \pi_{\theta}(a_t|s_t) \hat{A}_t(\theta) \right].
\end{align}
To improve the weights of the actor, the loss function is optimized through a \gls{sgd} method, which in case of the \gls{ppo} is the Adam algorithm \cite{kingma2014adam}. However, previous algorithms prior to the \gls{ppo} had  often  instabilities in their learning process, due to large updates in their parameters $\theta$. As a consequence, \cite{schulman2017proximal} address this problem by introducing $\epsilon$-clipping in their loss function to limit the update of the optimization:
{\small
\begin{align}
\begin{aligned}
    L_{t}^{CLIP}(\theta)\ =& \hat{E}_t\left[ \min(\tau_t(\theta)\hat{A}_t(\theta), \Gamma_{clip})\right],\\[12pt]
    \text{with} \  \tau_t(\theta)=&  \frac{\pi_{\theta}(a_t | s_t)}{\pi_{\theta_{old}}(a_t | s_t)},\\[12pt]
    \text{and} \ \Gamma_{clip}\ =&clip(\tau_t(\theta),1-\epsilon,1+\epsilon)\hat{A}_t(\theta) .
\end{aligned}
\end{align}
}
As one can see, the loss function is the estimated expectation of the minimum value of either $ \tau_t(\theta)\hat{A}_t(\theta)$ or the clipped value $\Gamma_{clip}$.\footnote{The $\tau_t(\theta)$ is the ratio between the old policy $\pi_{\theta_{old}}$ and the current updated one $\pi_{\theta}$.}  Very large positive updates in policy are clipped while negative updates are still considered valid. Thus, it is ensured that the updates are not running to a local maximum. Next to the optimization of the loss function, \cite{schulman2017proximal} further include other algorithmic advantages that enhance the optimization from a computational perspective. These are the parallel computation of multiple \textit{Actors}, the introduction of mini-batches and the multiple runs of \gls{sgd} on the same data. All of the advances increase learning speed and exploration of the environment as well as learning stability, thus making the \gls{ppo} our preferred algorithm. 

\subsection{Hyperparameter Optimization}
The \gls{ppo} algorithm has various hyperparameters, which have considerable influence on the training performance. Therefore, we included a hyperparameter optimization in our analysis. Under consideration of the \gls{drl} characteristics, we decided to use the \gls{pbt} hyperparameter search algorithm, based on the paper of \cite{jaderberg2017population}. The \gls{pbt} is constructed with an evolutionary concept, where multiple variations of the algorithm are trained simultaneously and the most fitting candidate is selected. For this purpose, the hyperparameters of multiple \gls{drl} agents are randomly sampled. After a specific training period, the performance of the agents is evaluated and the better performing candidates can continue their training. For the underachieving agents, the hyperparameters are re-sampled and in addition the model states of the better candidates are copied. Therefore, no training progress is lost, while simultaneously different options are tested.

\section{Research Design}
\label{sec:framework}
% Structure: 
% 1. Outline of research design:
\subsection{General Research Setting}
The trading environment is of huge importance for the overall success of the \gls{drl} algorithm and varies in the literature depending on the application of the agent. 
In our work, we construct research design from the perspective of a wind park operator, as stated in Section \ref{ssec:motivation}.
As a consequence, the task of the trading agent is to react to the changes in the wind forecast of the producer and automatically 
conduct correction processes on the \gls{id} market. For this purpose, we combine our trading environment with external wind forecasts to train the agent.
We evaluate our research on the German continuous \gls{id} electricity market, given that this market is frequently used in the literature, thus ensures comparability. Consequently, we shortly describe the structure of the continuous German \gls{id} market, derive restrictions and thereafter define our trading environment.

\subsection{The German Continuous Intraday Electricity Market}
\label{ssec:market} 
\begin{figure*}[ht]
    \centering
    \includegraphics[trim={2cm 9cm 3cm 0cm}, clip, scale=0.5]{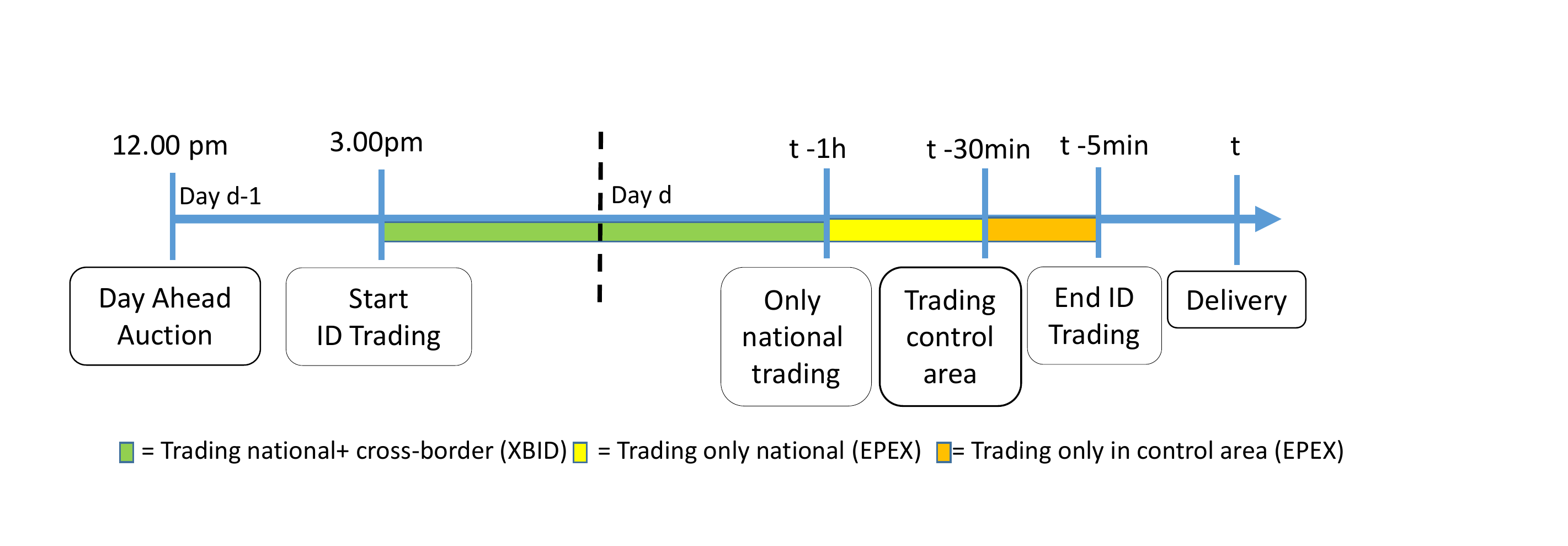}
    \caption{Overview over the trading stages of an \gls{id}  product with delivery at day $d$ and time $t$. The trading starts at 3.00pm on the previous day $d-1$ and consist of national and cross-border trading. One hour \gls{ptd} the cross-border trading stops, leaving national trading, and half an hour only trading in the control area is allowed.
    }
    \label{fig:tradingtime}
\end{figure*}

The German \gls{id} short-term electricity trade is primarily conducted on the EPEX SPOT exchange, where different electricity products, e.g., quarterly and hourly products, are traded by the market participants in a continuous manner. In this research we solely focus on the hourly products, which we hereinafter just referred to as products.\footnote{In future work, we plan to include the quarterly hour products in our analysis, especially because they are highly correlated to the hourly products.} The \gls{id} trade is based on the M7 trading system, thus the trade is executed through an individual \gls{lob} for each product, see \cite{martin2018german}. In each \gls{lob}, open buy/sell orders are collected, sorted according to their highest/lowest price and matched if a new order with a lower sell/higher buy price is submitted. The matched orders are then recorded in the transaction data.\footnote{Note that the trading process is heavily simplified, given that there are various order types and restrictions (iceberg orders, fill or kill orders, etc.). Further, there is also a chronological order to execute similar orders. For a detailed analysis, we refer to \cite{martin2018german} for more information.}
 For the hourly \gls{id} products, the trading interval starts at 3.00pm of the previous day and closes 5min \gls{ptd}, as seen in Figure \ref{fig:tradingtime}. While the first hours of trade are open for both national and cross-border trade (trade on the XBID exchange), the trading is gradually restricted in conjunction with the distance to the delivery time. At one hour \gls{ptd} only national trades are allowed, while in the last 30min until 5min \gls{ptd} only trading in the control area is allowed. As a consequence, the trading behavior differs, depending on the delivery time and the respective market participants. 

\subsection{Restriction for the Trading Environment}
\label{ssec:restriction}
As a consequence of the \gls{id} trade structure, it was necessary to construct an environment for the \gls{drl} agent capturing both the essential information of the \gls{lob}, while simultaneously reducing their complexity. Thus, multiple restrictions were necessary to model an adequate simulation of the trades. 

\begin{figure*}[ht]
    \centering
    \includegraphics[trim={0cm 0cm 0cm 0.8cm}, clip,scale=0.7]{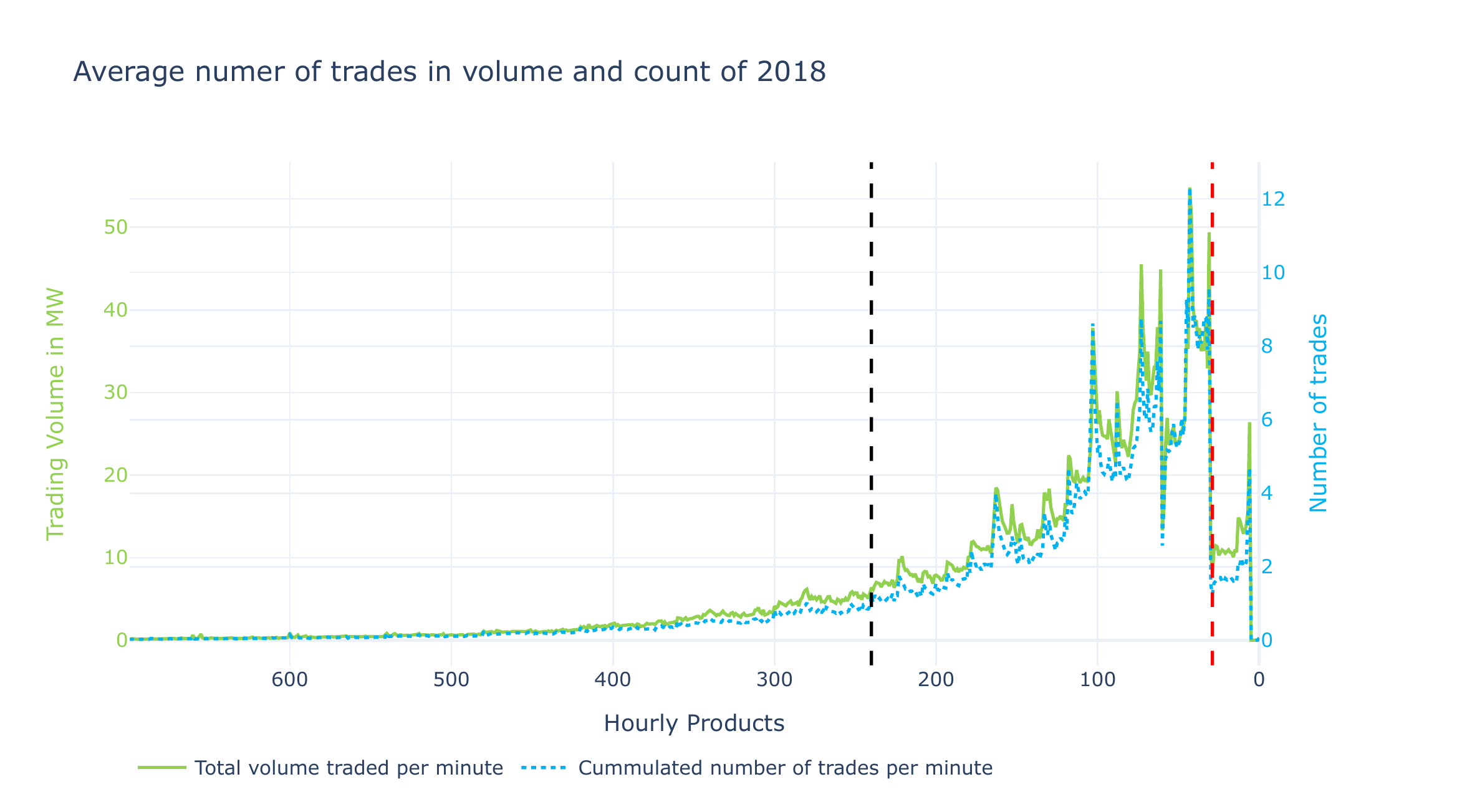}
    \caption{Visualization of the average number of trades per product, aggregated to minutely values, c.f. \cite{scholz2020towards}. The left axis depicts the aggregated volume in MWh, while the right axis displays the number of trades in the minute. Regarding the case study, the left vertical line (black) shows the beginning of our trading interval, while the right vertical line (red) shows the end of the trading interval.}
    \label{fig:num_trades}
\end{figure*}

First, we decided to build our environment only with the transaction data of the \gls{id} market, which is in most cases close to the actual \gls{lob}. This has the advantage that the transaction data consists only of traded contracts, hence excludes the various cases of open orders, trade limitations, over the counter contracts and other redundant information. One consequence arising with this restriction is that we establish our agent as a price taker with no direct influence on the \gls{id} price. Under the assumption that the agent is only trading small amounts 
of energy, this seems reasonable. If one wants to trade a larger amount of electricity, multiple orders from the \gls{lob} have to be considered, which might result in a disparity between the traded \gls{lob} price and the transaction prices. Second, while the electricity trade is continuous, i.e., traded in milliseconds, we limited our data to minutely frequency by aggregating one minute volume weighted averages of the price.% With the volume weighting, we diminished the influence of outliers in the transactions and ensure a more stable training of the agent. 
Third, we restricted the trading time of the agent to the last four hours until the last 30min \gls{ptd}. This step was justified, under the consideration that most of the trades were executed in the last hours \gls{ptd}, as seen in Figure \ref{fig:num_trades}. In addition, we had not sufficient information on the trade in the control areas, thus we excluded the last 30 minutes \gls{ptd} as well. %Further, we were only interested in national trading and therefore also excluded the last 30 minutes \gls{ptd}. %Moreover, under consideration that the overall trading time differs depending on the delivery hour, we were able to ensure similar length in the price time series. 
Lastly, the fourth restriction is solely for the training data, because we excluded products that consisted of extreme outliers. This step was necessary to ensure stable training performance, however for the testing scenario no exclusion was conducted. 

\subsection{Detailed Description of the Environment}
\label{ssec:env}
With the outlined restrictions, the trading environment $\mathbb{S}$ was constructed based on the methodology of Section \ref{ssec:mdp}. Hereby, the primary goal was to provide a trading environment that was close to the actual \gls{id} trade and incorporated most features. We followed the widely accepted Gym format of OpenAI \cite{OpenAI} and again refer to Figure \ref{fig:overview} for a better overview.

\paragraph{Episodes}To begin with, each hourly product was considered an individual episode in the environment. Depending on the scenario, the products were either randomly sampled (training) or processed sequential (testing). In the episode, the one minute volume weighted averages of the transaction price was computed. However, at some minutes no new observations were available, i.e., no new trade occurred, resulting in a forward passing of the previous price in order to ensure minutely coverage. As explained in the previous section, we fixed the number of trading steps to a length of 3.5 hours, thus each episode had a total of 211 time steps. Note that the agent had no influence on the completion of the episode, which differs from other \gls{drl} environments.

\paragraph{Action Space} For the action space $\mathbb{A}$ of the environment, we established a continuous action space, which should reflect the traded volume of the wind turbine on the electricity market. We restricted the operation margin, by setting lower and upper limits of MWh to our trading interval $a_t\ \in \ [0,1]$ for all $t \in T$. At each time step $t$, the agent adjusted through its action $a_t$ the traded volume and the difference $\Delta v_t = a_t-a_{t-1}$ thereafter was traded with the respective electricity price.\footnote{As example, the current traded portfolio of wind power consist of 0.5MWh for a specific product at time step $t=3$. The \gls{drl} agent proposes the action $a_4 = 0.6$, thus the traded volume has to be increased by $v_4=0.1$, i.e. 0.1MWh have to be sold on the \gls{id} market.}  %Even though the average capacity of a wind turbine (onshore) is around 3.4 MWh in Germany \cite{windstat}, 
With regard to the trading interval $a_t$, we decided to restrict the action for the following reasons. 
First, through only trading a maximum of one MWh, we could ensure that the price of the transaction was relatively similar to the \gls{lob}, as outlined in the previous section. Second, the interval simplified the interpretation of the results. Third, the forecasted wind volumes were also normalized within the interval [0,1], therefore the limit of $\mathbb{A}$ ensured a straightforward implementation.

\paragraph{Observation Space} The observation space of the \gls{drl} agent, i.e., the representation of the current state $s_t$, consisted of twelve features. We list the full table of our observation variables in Table \ref{tab:obs_space} in the \ref{apx:obsspace}. Next to the current price of the \gls{id} market $p_t$ and the forecast of the production volume $\eta_t$ (Section \ref{p:windfcst}), ten additional features were included to increase the performance of the agent.  These features can be differentiated in \textit{price variables}, \textit{volume variables} and \textit{episode variables} and the most important ones are described below.  The first important feature is the 5min-forecast of the electricity price $\hat{p}^{5min}_t$ (Section \ref{p:pricefcst}), which we introduce to indicate possible future trends for the agent. Further, we calculated the value of the portfolio by averaging the selling price in relation to the respective volume, denoted as $\tilde{p}^{port}_t$. 
With regard to the volume related variables, we included the current volume sold to the market, i.e., the previous action $a_{t-1}$ to ensure that the agent knows the available quantity for trading. In addition, we also included a difference between the current volume and the predicted production of the wind forecast, $vol_{diff}= \eta_t - a_{t-1}$. Lastly, we also added a distance measure that denotes the time until the trade ends $tte$, given that the variance of the \gls{id} price increases in the last minutes of the trade.

\paragraph{Reward} From the producer point of view, an increase in the trade volume would correspond to selling more capacity on the market and respectively a decrease in volume would correspond to buying back some electricity. Accordingly, we reflected this relation in the reward $r_t$ of the agent, which we defined for each time step $t$ as follows: 
\begin{align}
    r_t &= r_t^{vol} + r_t^{trade}. 
\end{align}
As one can see, the reward was separated into two parts, with the first part being the trade reward $r_t^{trade}$ and the second the volume reward $r_t^{vol}$. The trade reward is valid for all time steps $t$ of the episode and is defined as follows:
\begin{align}
\label{equ:reward}
\begin{aligned}
       r_t^{trade} &= p_t \Delta v_t-c_{fee} \mid\Delta v_t \mid, \\[5pt]
        \text{with} \ & \Delta v_t = a_t-a_{t-1}
\end{aligned}
\end{align}
As one can see, $r_t^{trade}$ the denotes the selling/buying of the action difference $ \Delta v_t$ to the current price $p_t$.  Further, we also included a penalty for trading which reflects the transaction cost $c_{fee}$ and was set to 0.2\euro{}/MWh after consultation with industrial experts
Thus, if $p_t>0.2$ an increase in volume (sold MWh) corresponds to a positive trade reward, while a decrease in volume (buy MWh) corresponds to a negative trade reward. Regarding the second part of the reward, the $r_t^{vol}$ is only computed in the last step of the episode $T$ and is defined as follows: 
\begin{align}
\begin{aligned}
        r_t^{vol} &= \left\{\begin{array}{lr}
        0 &\ \text{for } t< T\\
        -0.1\Delta v_{penalty}^2 \ &\  \text{for } t= T
         \end{array}\right. ,\\[12pt]
        \text{with }&
        v_{penalty}=\eta_{T}-a_{T}.
\end{aligned}
\end{align}
The $r_t^{vol}$ is a penalty reward that measures the distance between the last action of the agent $a_{T}$ and the last provided wind forecast $\eta_{T}$. Through the quadratic term, the volume reward penalized large deviations, thus ensuring adaption at the end of the episode.

\subsection{External Forecasts}
In order to ensure a realistic trade behavior of the agent, we provided two forecasts in the observation space. 
The forecasts in question were on the one hand the wind power forecast of the wind farm $\eta_t$ and on the other hand a forecast of the next five minute price average $\hat{p}^{5min}_t$. In the following, we outline their structure shortly.\footnote{The hyperparameter of the forecast models are in the \ref{apx:Hyperparam}.}

\paragraph{Intraday Price Forecast}
\label{p:pricefcst}
The \gls{id} price forecast was constructed similar to the paper of \cite{scholz2020towards}, where both a \gls{lstm} and XGBoost model were used to predict future prices. Similar to \cite{scholz2020towards}, we used the XGBoost model for the forecast. However, contrary to cite \cite{scholz2020towards} we decided to forecast the weighted \gls{id} price in 5min intervals, instead of 15min intervals. This has the advantage that short term changes are anticipated faster, however sacrificing at the cost of loosing mid-range information. In order to predict a specific month, the previous three months are considered as training interval. Note that we retrained the models for each new month to embed seasonal differences in the models.
With regard to the input data, we included the transaction and \gls{lob} data of the EPEX SPOT market as well as the national wind forecast error of Germany. 

\paragraph{Wind Forecast}
\label{p:windfcst}
In terms of the wind power forecast, we wanted to ensure a realistic setting by using a current state-of-the-art wind power prediction model. Therefore, we implemented an encoder-decoder \gls{rnn} \cite{sutskever2014sequence} for a multi-step-ahead forecast of the power generated by a single wind turbine. Encoder-decoder \gls{rnn} consist of an encoder network that processes an input sequence and encodes it into a latent representation. The decoder network thereafter translates the encoded information to sequentially generate the output sequence, with each previously generated element influencing the next. Encoder-decoder \gls{rnn} have been used several times in the literature for wind and solar power prediction \cite{alkhayat2021review}. In case of our analysis, we used an encoder consisting of an \gls{lstm} layer with 50 neurons. As a decoder, we used an \gls{lstm} layer with 50 neurons, followed by a dense layer with one neuron and a leaky ReLU activation function. The encoder received wind speed forecasts and wind power measurements from the past 28 hours as input. The decoder was initialized with the encoder's hidden states. Further it received the wind power prediction of the previous prediction step such as the wind speed forecast of the current prediction step as input. The wind speed forecasts for the location of the wind turbine were extracted from the weather forecasts of the ICON-EU model, provided by the German Meteorological Service \cite{dwd}. 

\section{Case Study}
\label{sec:experiment}
\subsection{Electricity Trading on the German ID Market of 2018}
\begin{figure*}[ht]
    \centering
    \includegraphics[ trim={0cm 0cm 0cm 0.8cm}, clip,scale=0.7]{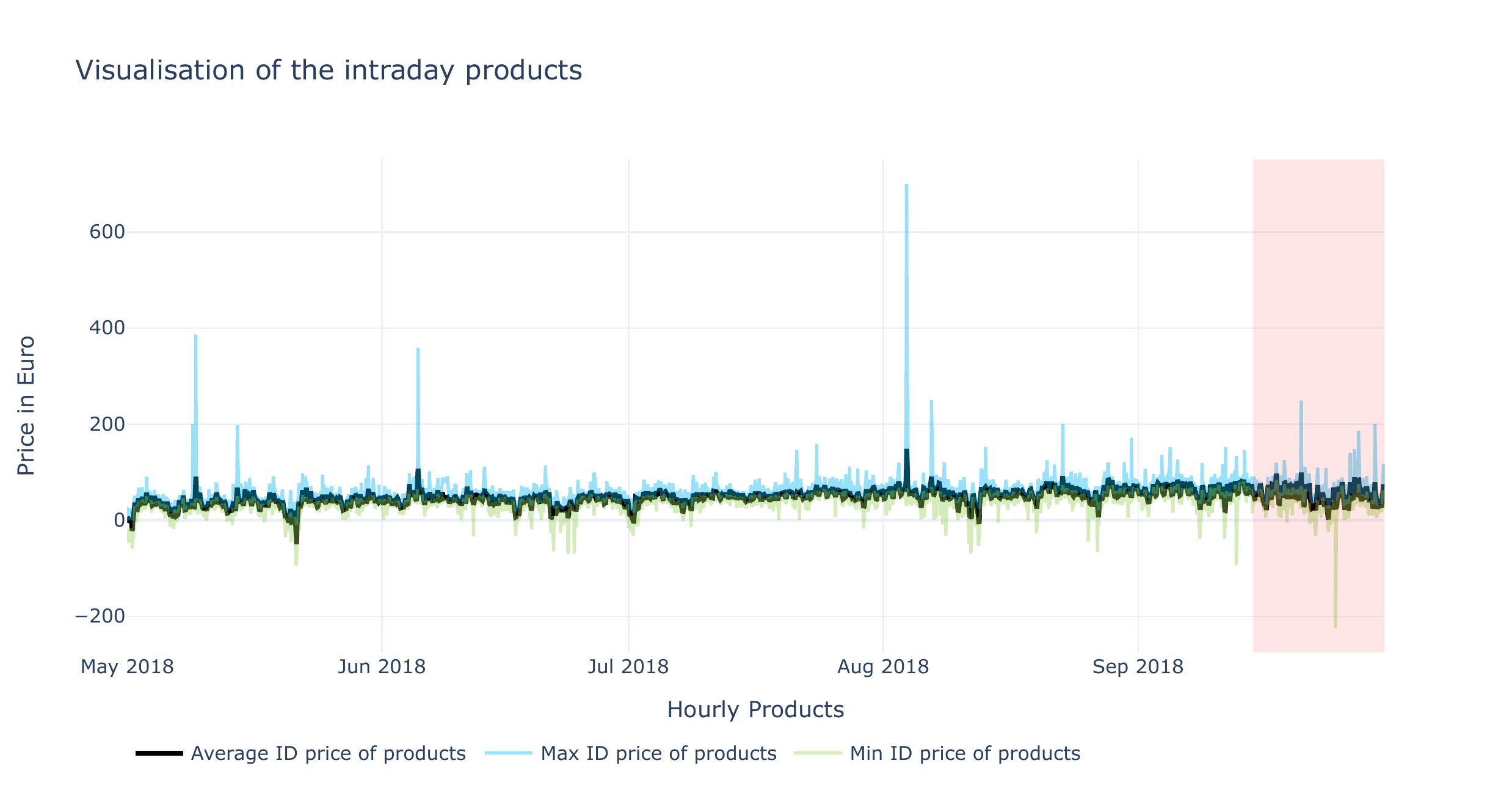}
    \caption{Plot of the 2018 German electricity \gls{id} price. Each product is visualized with its average, maximum and minimum value of the \gls{id} price. The training set is from May 2018 till mid of September, while the testing set is the remaining September (red).}
    \label{fig:spotprice}
\end{figure*}

In this case study, we analyze the behavior of the \gls{drl} trading agent based on historic trading data from the German \gls{id} market. As primary data set the EPEX SPOT \gls{id} transaction data of 2018 was used for the trading environment, because the data was already available to the institute. With regard to trading regulations, we want to point out two market changes that happened in 2018. First, in June 2018 the XBID trading went live, which enables \gls{id} cross border trading \cite{press1epex}. Second, the German-Austrian bidding zone was separated in October 2018 into two different market areas \cite{press2epex}. Considering the impact of the first change, \cite{kath2019modeling} found that the introduction of the XBID had no direct influence on the trading behavior, thus could be disregarded for market modeling. However, in terms of the market split we could not find similar analysis, thus decided to only selected the summer months of 2018 as our data set. In order to ensure impartial results, we proposed 10\% of our data as out-of-sample test data, resulting in a training period from the 01.05.2018 till the 14.09.2018, with the remaining data 15.09-30.09.2018 as independent testing interval. The test interval represents the overall \gls{id} price of 2018 sufficiently and incorporate both products with average price distributions as well as outliers in the \gls{id} price. We present a more detailed analysis and comparison in the \ref{apx:Testsample}. 
Further, to give a better overview, we display for each product the average \gls{id} price as well as the minimal/maximal outlier in Figure \ref{fig:spotprice}, with the testing interval underlined in red. As one can see, there are some extreme outliers in the data that might result in unstable training  behavior. Consequently, training products with values above 150\euro{} or below -50\euro{} were discarded, which totals in 33 unregarded products. For the testing interval no products were filtered, given that these observations would not be known beforehand. Overall, we have in total 3.255 training and 384 testing products, which translates into 686.805 training and 81024 testing observation in our analysis. 

\subsection{Baseline Agents}
\label{ssec:BL}
As part of the case study, we want the agent to perform against different baseline agents to have an objective benchmark and asses the performance of the 
\gls{drl} agent correctly. %Through the analysis of the different agents, we can verify the capability of the \gls{drl} agent and analyze possible improvements.
However, some ground rules are necessary to ensure equal chances. First,  all agents, including the \gls{drl} agent,  had to start with a trading volume of 0 and could only trade in between the interval of [0,1] MWh. Further, to ensure physical delivery all agents were forced to end their trading on the last value of the wind forecast, which we corrected automatically at the end of each episode. Through fixing the start and end volume of the trading interval, we ensured a fair comparison, given that the performance was only depending on the movements within the trading interval.

With respect to the individual baseline agents, different approaches were chosen that should reflect different perspectives of the energy trading. The first and easiest agent $BL_{\text{First}}$ sold in the first step the volume according to the first wind forecasted volume $\eta_0$ at time $t_0$ without any other considerations. Then in the last step, it corrected the volume to the last predicted value of $\eta_T$. In conceptual terms, the second agent $BL_{\text{WF}}$ was constructed  similar, because it strictly followed the wind volume forecast and traded every change (every 15 minutes) immediately. This agent therefore corresponds to a very conservative rule-based trading approach. 
The remaining agents are more sophisticated to reflect more complex approaches. The third agent $BL_{\text{PF}}$ is based on the \gls{id} price forecast values $\hat{p}^{5min}_t$ and should capture potential arbitrage possibilities.
If for two sequential periods the difference between the price and the forecast indicated a smaller/larger energy price, the agents sold/bought a fixed proportion of its volume (0.1MWh) on the energy market. Our reasoning for proportional trading is that a higher proportion also induces higher transaction cost. Further, regarding the two sequential periods, we argue that the forecast $\hat{p}^{5min}_t$ is an average forecast for the next five minutes and hence captures a trend that might occur in later time steps. These constrains were experimentally found and showed better results than simply trading all capacities whenever the difference between forecast and price appeared. 
The last agent $BL_{\text{Random}}$ is a stochastic approach that should simulate a random behaviour on the \gls{id} market. With a probability of 25\%, the baseline samples a new trading volume from a normal distribution. Thereby, the mean was the current wind forecast for the product, while the standard deviation was the standard deviation of the training data set. 
The 25\% probability ensures that the agents performance is not to much influenced by the transaction costs of 0.2 \euro{}/MWh .

\subsection{Experiment Metrics}
\label{ssec:metric}
For the comparison of the agents, we use multiple experiment metrics which we shortly introduce. First, we calculate the \textit{overall profit}, which is the cumulative return across all test products, subtracted by the transaction costs (0.2\euro{}/MWh). Second, we derive a percentage performance in comparison to cumulative return of the $BL_{\text{WF}}$ baseline. Third, we present distributional properties with the \textit{mean}, the \textit{standard deviation} as well as the  \textit{10\% and 90\%Quantiles} to give a better understanding of the allocation of the trading results. In addition, we were also interested in the best trade result, which the respective agent had in comparison to its peers. Therefore we return a  relative measure in \% in the results. Lastly, we conduct the average number of actions per product for the respective agent. 

\section{Results}
\label{sec:results}
\subsection{Trading Results}
\label{ssec:trade_results}

\begin{figure*}[tp]
\centering
\begin{subfigure}{1.0\textwidth}
  \centering
    \includegraphics[ scale=0.65]{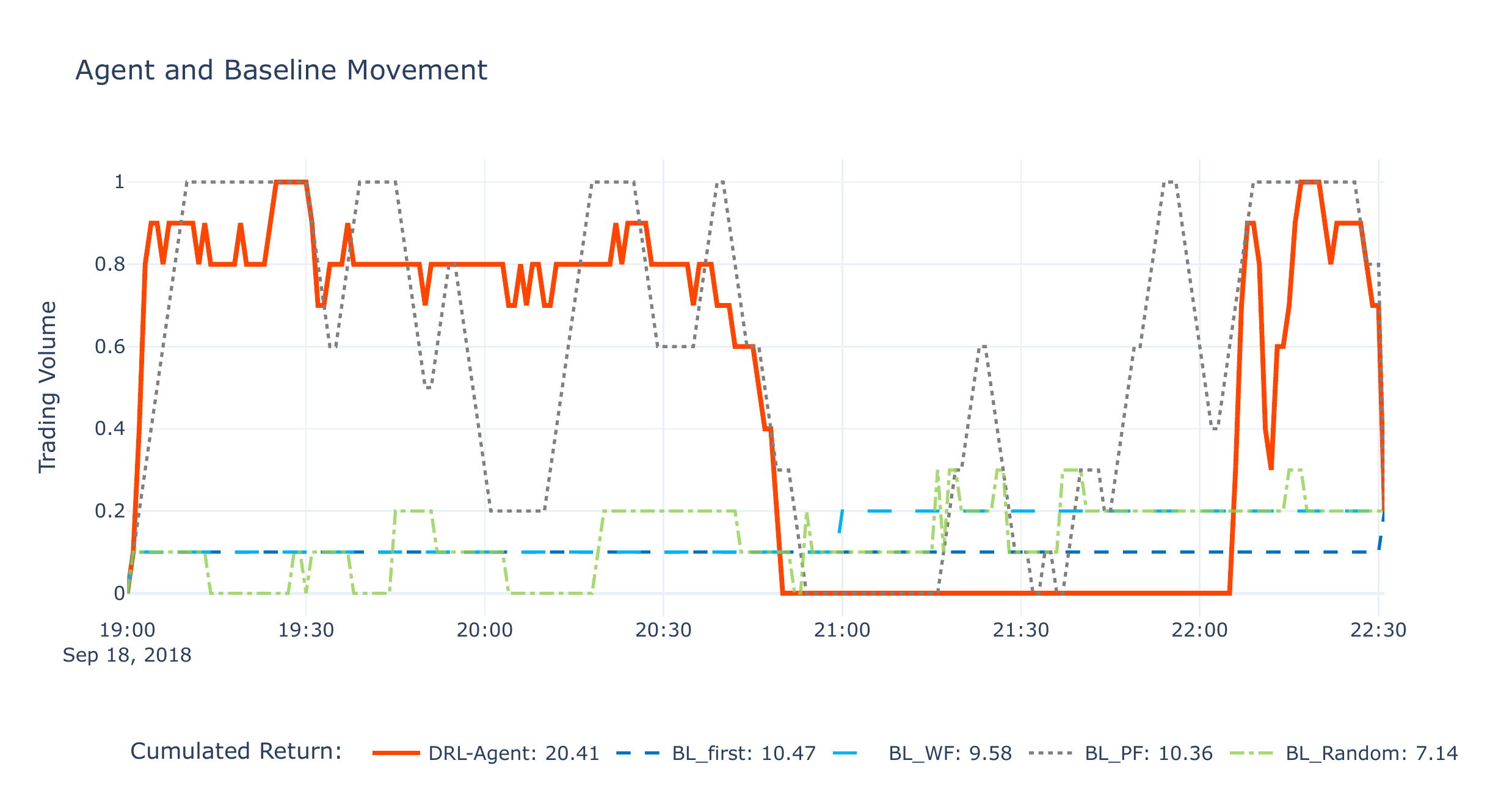}
     \caption[Short Title]{
     Visualization of the baseline and \gls{drl} agent actions for the product of 2018-09-18 at 23.00. The trading period starts at 19.00 with a volume of 0 MWh and ends at 22.30 with a volume of 0.2MWh.}
  \label{fig:movementagents}
\end{subfigure}
\begin{subfigure}{1.0\textwidth}
  \centering
    \includegraphics[scale=0.7]{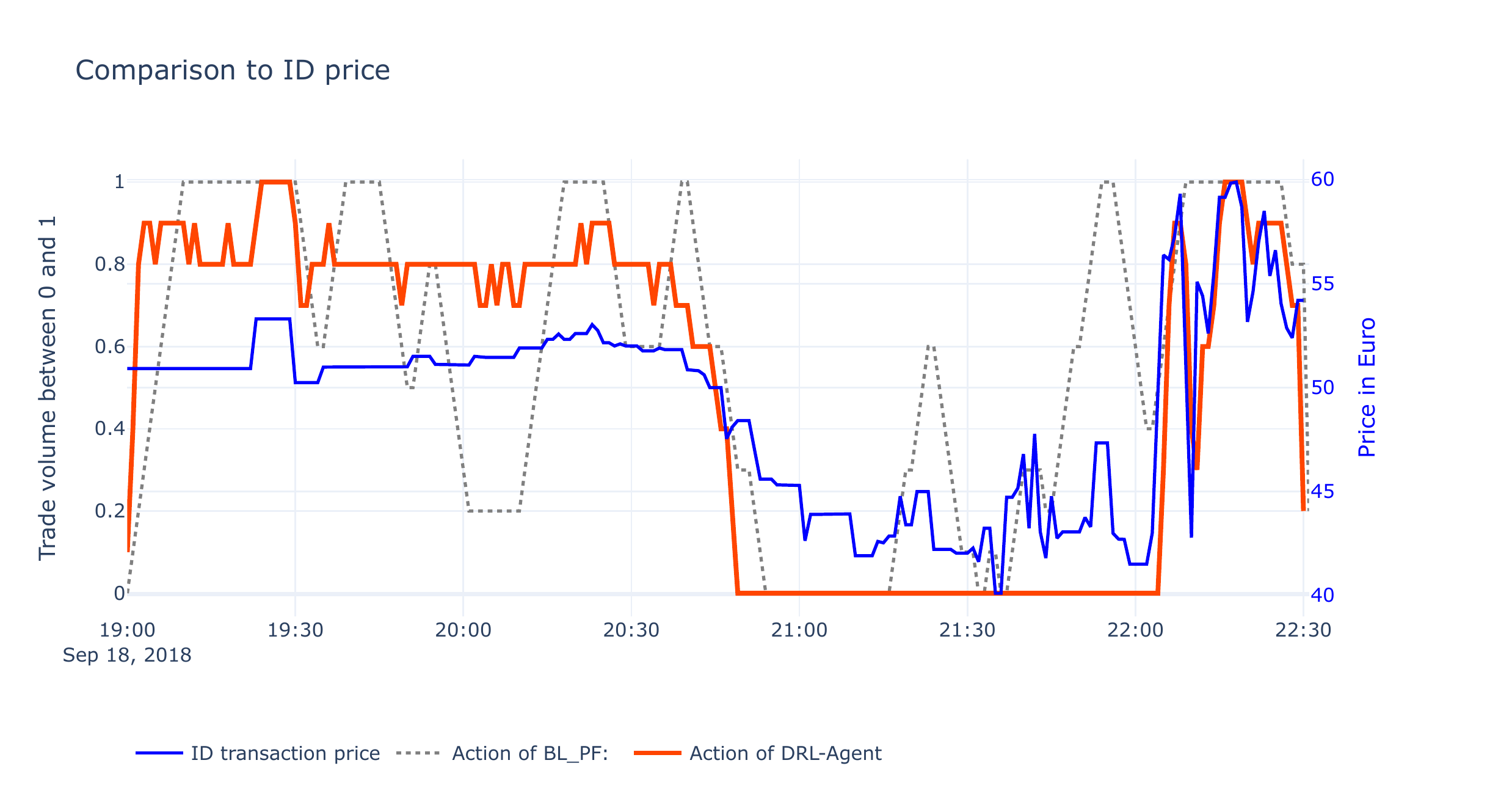}
    \caption[Short Title]{This plot shows the actions of the \gls{drl} agent and $BL_{\text{PF}}$ in comparison to the \gls{id} price of the 2018-09-18 at 23.00 product. The left y-axis displays the trading volumes, while the right y-axis corresponds to the \gls{id} price. }
  \label{fig:pricemove}
\end{subfigure}
\caption{Visualization of agent and baseline performance on the  exemplary product 2018-09-18 at 23.00.}
\label{fig:moveit}
\end{figure*}
Based on the specifications of the case study, the results are shown in the following section. To begin with, we first present an exemplary product and thereafter analyze the overall results of the agents.  

\paragraph{Exemplary Product}
We selected an exemplary product 18.09.2018 at 23:00 UTC to visualize the actions of both agent and the baseline methods.\footnote{For the selection, we sampled from the test products with the condition that at all agents executed at least one trade movement.} In Figure \ref{fig:moveit} the respective actions are plotted, while in Figure \ref{fig:pricemove} we plotted the price history of the product and compare it to the agent and $BL_{\text{PF}}$ action. 
Beginning with the baselines, one can see that both the $BL_{\text{First}}$ and $BL_{\text{WF}}$ were only trading twice, due to the minimal change in the wind forecast. Similarly, the $BL_{\text{Random}}$ had a relative narrow range in its movement, resulting in only a small deviation from the forecast. Consequently, all three agents had similar trade results.% with 10.47\euro{}, 9.58 \euro{} and 10.19 \euro{} respectively. 
In contrast, the \gls{drl}-agent and $BL_{\text{PF}}$ had large changes in their trading action and use the full potential of their trading range. Both agents first sold all their available capacity in the first 30 minutes, re-bought when the price dropped (at around 20.45) and then again sold their capacity when the price was rising again. However, after 21.00 the $BL_{\text{PF}}$ started to sell parts of their capacity, while the agent seemed to anticipate future price trends and only moved after 22.05. Consequently, when looking at the Figure \ref{fig:pricemove} one can see that the agent was exploiting the price drops at a lower price level than the $BL_{\text{PF}}$. Further, we want to draw the attention to the spike of the agent at 22:11, where the agent exploited the sudden plunge of the \gls{id} price. As a result, the agent was able to achieve a total return of 20.41\euro{}, whereas the $BL_{\text{PF}}$ only managed a total of 10.36\euro{}. 

\begin{figure*}[ht]
    \centering
    \includegraphics[trim={0.5cm 0cm 0cm 0.8cm}, scale=0.7]{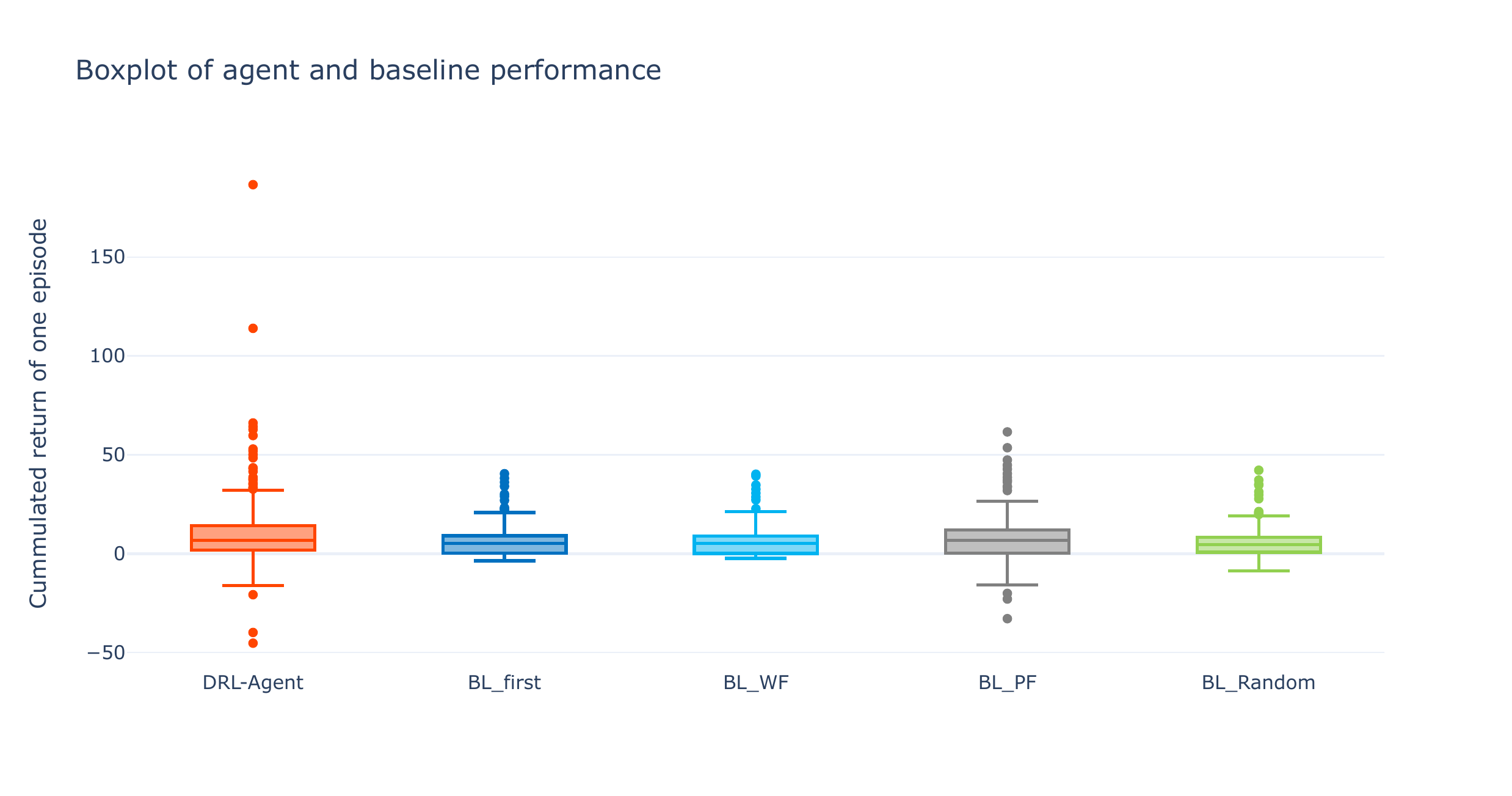}
    \caption[Short Title]{Box-plot of the performance from \gls{drl} agent and the baselines. The y-axis denotes the total return from each product of the test sample.}
    \label{fig:boxplot}
\end{figure*}

\paragraph{Overall Performance}
While the visualization of a single product offers a first idea on the typical action range, we are naturally more interested in the overall trading results. Therefore, we visualize the performances across all products through a boxplot in Figure \ref{fig:boxplot}. 
The first thing that becomes apparent is that both the \gls{drl} agent as well as the $BL_{\text{PF}}$ had a larger spread in their performance in comparison to the other more narrowly constructed baselines. This spread does not only correspond to higher trading values but also includes more negative and positive trading results. On the one hand, the \gls{drl} agent traded twice with a negative reward of -39.86\euro{} and -45.27\euro{}. On the other hand, the agent was also able to achieve extraordinary returns with a maximum returns of 186.69\euro{} and 114.00\euro{}.\\
\begin{table*} [ht]
\begin{tabular}{l|rrrrr}
\toprule
{} &    Agent &  $BL_{\text{First}}$ &  $BL_{\text{WF}}$ &  $BL_{\text{PF}}$ &  $BL_{\text{Random}}$ \\
\midrule
Mean                        &     9.72\euro{} &           6.23\euro{} &        6.15\euro{} &         6.69\euro{} &        5.50\euro{} \\
Median                      &     6.86\euro{} &           5.24\euro{} &        5.26\euro{} &         6.84\euro{} &        4.59\euro{} \\
Standard Deviation          &    16.53\euro{} &           6.42\euro{} &        6.18\euro{} &        10.47\euro{} &        6.42\euro{} \\
10\% Quantile                &    -1.01\euro{} &           0.00\euro{} &        0.00\euro{} &        -4.73\euro{} &       -0.79\euro{} \\
90\% Quantile                &    23.88\euro{} &          13.63\euro{} &       13.45\euro{} &        16.45\euro{} &       12.65\euro{} \\
\hline
Total net profit            &  3731.69\euro{} &        2390.85\euro{} &     2360.56\euro{} &      2569.22\euro{} &     2111.08\euro{} \\
\% Improvement to $BL_{WF}$ &    58.08\% &           1.28\% &        0.00\% &         8.84\% &      -10.57\% \\
\hline
Best Performance in \%       &    40.24\% &          15.48\% &       11.90\% &        28.57\% &        3.81\% \\
Steps                       &    27.95 &           1.08 &        1.34 &        77.89 &       26.59 \\
\bottomrule
\end{tabular}
\caption{Trade results of the \gls{drl} agent and the baseline agents based on the performance metrics of Section \ref{ssec:metric}. }
\label{tab:result}
\end{table*}

Given that the one can not fully distinct between the median and quantiles of the boxplot, we report the performance metrics in the Table \ref{tab:result}. Here, we can see that the \gls{drl} agent was in both mean and median better than the baselines, with the $BL_{\text{PF}}$ at second place, followed by the $BL_{\text{WF}}$, $BL_{\text{First}}$. The worst performance was the $BL_{\text{Random}}$, which was not surprising and shows that random behaviour does return sufficient results on the \gls{id} trade.
Note that regarding the medians, the median of $BL_{\text{PF}}$ and the agent were quite similar. 
In addition, we are able to confirm the assumption that the agent and $BL_{\text{PF}}$ had a larger variance than the remaining baselines, which is both visible in the standard deviation as well as the quantiles. 
Naturally, when analysing trading methods one is primarily interested in the total net profit, where the agent was able to achieve the highest return with 3731.69\euro{} and outperformed the $BL_{\text{WF}}$ by 58.08\%. With comparison to the best baseline ($BL_{\text{PF}}$) this results to a surplus of 1162,47\euro{}. Considering the best performance per product, we see that the agent was able to achieve in 40.24\% of times the best performance in direct comparison to the other baselines. Interestingly, the $BL_{\text{PF}}$ was following relatively close with 28.57\%, while the remaining agents were clearly exceeded. 
Lastly, we consider the average number of steps per product, where the agent did an average of 27.95. In comparison to the 77.89 ($BL_{\text{PF}}$) and 26.59 ($BL_{\text{Random}}$) steps, these results conform the visual examination of the example product, where the agent had both some active trading intervals but also some times, where the action did not change. 

\subsection{Discussion and Future Outlook}
\label{ssec:discussion}
Under consideration that this paper proposed a first attempt at a shortest-term trading  \gls{drl} agent, we were able to show that it was indeed possible to apply \gls{drl} on the \gls{id} electricity trade in a minutely resolution. Even though the \gls{id} price emits a very high variance, the agent was able to identify patterns and create a sufficient trading strategy. This was primarily achieved through the development of the \gls{drl} environment. By excluding extreme outliers and implementing the necessary restrictions, we were able to create a training environment, which returned sufficient performance. %The \gls{pbt} was also crucial, due to the fact that it shaped the hyperparameters depending on the training progress. 
The overall results clearly indicate an advantage through the \gls{drl} approach, considering that the agent outperformed all baselines, with best performances in 40.24\% of the products. However, in comparison to the simple forecast of the $BL_{WF}$ and $BL_{first}$, we have to point out the higher variance in the trade results of the \gls{drl} agent. When looking at the highest negative return of the agent, i.e., the 2018-09-20 20:00:00 product with a -45.27\euro{} loss, we observed that this product had large price increase up to 176.61\euro{}. The agent was not able to detect the price increase, which might partly be explained by the exclusion of extreme outliers from the training data. This might be a drawback for investors that are risk avers and do not want to anticipate negative results. Therefore, in future work one has to find a solution to cope with extreme outliers to ensure more stable results. Another deduction we have to make, is the role of the short-term price forecast. Its inclusion showed a clear increase in the agents performance and also returned better results in the $BL_{PF}$ baseline. Regarding the median results of the agent and $BL_{PF}$, there might be an indication that the average products can indeed be traded with the price forecast. However, for more volatile products the \gls{drl} agent provided a better trading strategy, resulting in overall better performance. Nevertheless, it will be interesting in future analysis, whether a more complex rule-based approach is able to achieve similar results than the agent. Lastly, given the limitations of our data, we want to address the fact that we evaluated the agent only on test data from one time period. On the one hand, the sample was a relative sufficient representation of the \gls{id} price from 2018, given that products with various price distributions 
were included, again see \ref{apx:Testsample}. On the other hand, a two week period might not fully reflect possible long term influence, e.g., seasonal effects, that have an impact on the \gls{id} price. Therefore, to ensure reliable results, it will be necessary in future work to test the agent on large test samples as well as newer data.  In this context, it might be interesting to enhance the performance through the inclusion of additional variables. One possible candidate could be the order book data, both as input variable as well as in the price mechanism. This would increase the complexity of the environment,
considering that the bidding with orders and taking completed orders of the market needs to be included in the environment.

%Regarding the future outlook, there are various ways to increase the difficulty and the performance of the agents. To begin with, the presented environment will be enhanced with newer data and also new variables.#
%However, this would also ensure a more realistic setting in general. 
%An additional extension could also be the increase in the frequency, thus analyzing the trade in seconds. Though this would also lead to even more randomness in the data, which the agent has to anticipate. 

\section{Conclusion}
\label{sec:conclusion}
In this work, trading on the \gls{id} electricity market was formulated as a  \gls{mdp} in order to apply the \gls{drl} algorithms for autonomous trading.
The novelty of the approach was the shortest-term resolution of one minute steps to depict the continuous trade. We offer multiple advancement to create a \gls{rl} environment in which an \gls{drl} agent can train, while simultaneously remaining a realistic setting. The \gls{drl} environment was tested in a case study from the perspective of an electricity provider to manage a wind park with data from 2018.  Our experiments show sufficient results with the agent outperforming the baselines by 45.24\%. 
We discuss the implication of the results as well as limitations of our first approach and further offer possible enhancements for future research.

\section*{Acknowledgement}
This work was supported by the Competence Center for Cognitive Energy Systems of the Fraunhofer IEE.

%\bibliographystyle{elsarticle-num}      % mathematics and physical sciences
%\bibliography{det_bib}


\newpage
\appendix
\label{sec:appendix}
\onecolumn

\begin{landscape}

\section{Input Variables of Observation Space}

\label{apx:obsspace}
In this section of the appendix, we present the underlying variables of the observation space to more detail. 
\begin{table}[h]
    \centering
    \begin{tabular}{l|l|c}
      Variable Group &Variables & Abbr. and Equation  \\[3pt]
      \hline
        \multirow{6}{*}{Price Variables}& Current and previous volume weighted electricity price & $p_t,p_{t-1}$  \\[3pt]
        & Day-ahead price & $P_{day_ahead}$  \\[3pt]
        &Forecast of electricity price & $\hat{p}^{5min}_t$ \\[5pt]
        & Current and previous difference between price and forecast & $d_t, d_{t-1} \text{ with } d_t = p_t  -\hat{p}^{5min}_t $ \\[3pt]
                &Portfolio price of the current volume & $\tilde{p}^{port}_t$ \\[5pt]
        & Price Marker & $m = \left\{\begin{array}{lr}
                                  1 &\ \text{for } d_t > 0 \text{and} d_{t-1} > 0\\
                                  -1 &\ \text{for } d_t < 0 \text{and} d_{t-1} < 0\\
                                0 & else
                                \end{array}\right.$  \\[3pt]
        \hline
        \multirow{4}{*}{Volume Variables} &  Forecast of the production volume from wind park &  $\eta_t$ \\[3pt]
        &Current volume of the trading agent & $a_{t-1}$ \\[3pt]
        &Difference of current volume and production volume & $vol_{diff}$ \\[3pt]
        \hline
        \multirow{1}{*}{Episode Variable} &Time to end & $tte =1 - \frac{t}{T}$ \\
        \hline
    \end{tabular}
    \caption{Overview over the observation space variables. All variables are normalized in order to ensure stable training results. Note that the day ahead variable $P_{day_ahead}$ is constant within each period and therefore has no time index. }
    \label{tab:obs_space}
\end{table}

\end{landscape}

\newpage
\section{Testsample}
\label{apx:Testsample}
\begin{figure*}[ht]
    \centering
    \includegraphics[trim={0cm 0cm 0cm 0.8cm}, clip,scale=0.8]{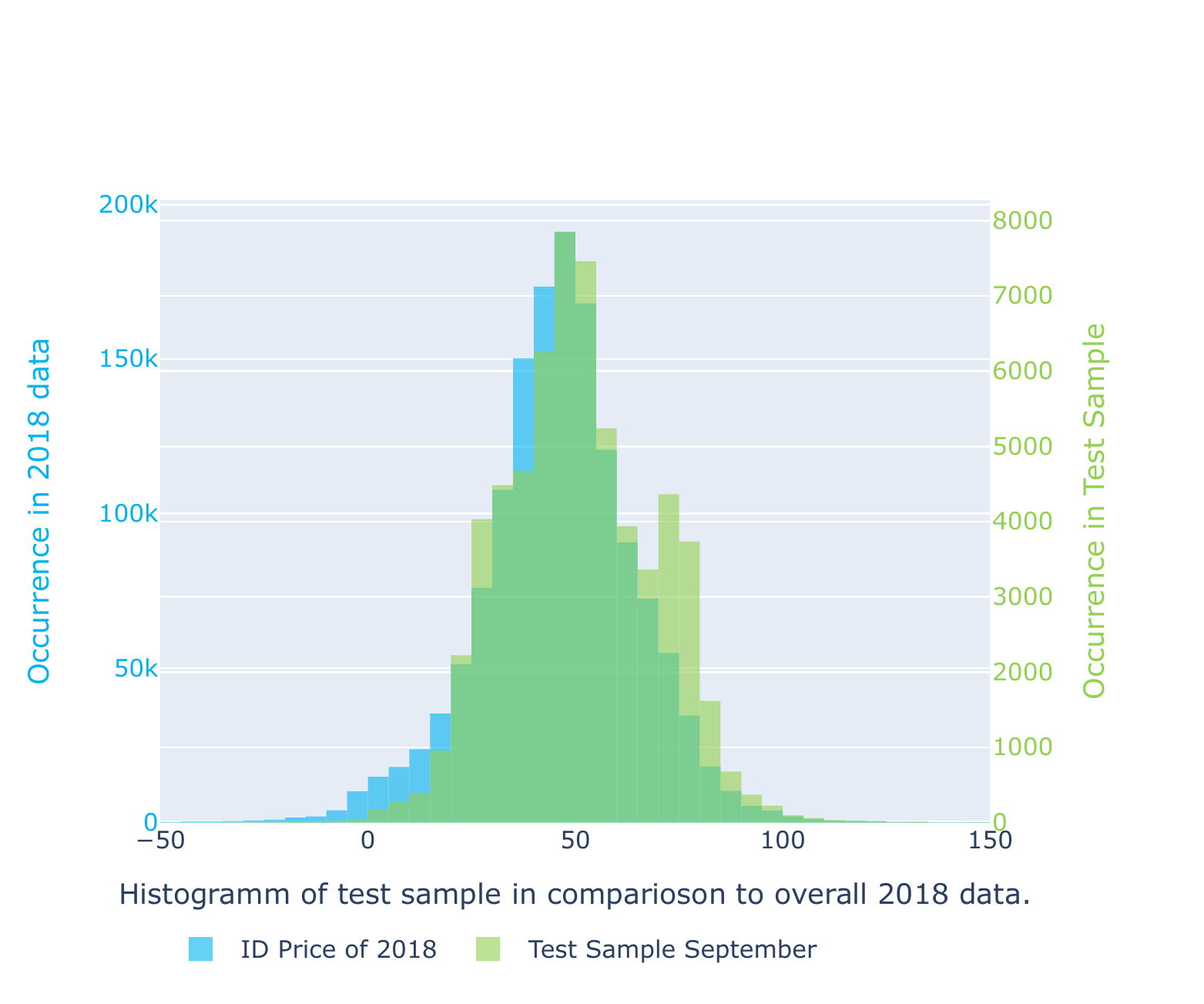}
    \caption{Histogram of the 2018 products in comparison to the products of the test sample. The overlap in the distribution can be seen as the dark green area in the histogram. Note that the different y-axis correspond to the frequency in the histogram-bins. }
    \label{fig:hist}
\end{figure*}

As a test sample, 10\% of the overall data was selected to validate the performance of the agents. We decided to use the products from 15.09.2018-30.09.2018 under consideration that these products represent the overall data sufficiently. In the test sample, the product 25.09.2018 records with -224.43\euro{} the lowest \gls{id} price of 2018. Additionally, the products 2018-09-20 20:00:00 and 2018-09-29 19:00:00 are with their maximum values of 200\euro{} and 249\euro{} within the top 20 of the highest products of 2018. Therefore, they might offer some insight in the outlier behavior of the model. The appropriateness of the test data set can also be seen in Figure \ref{fig:hist}, where we visualize the overall \gls{id} prices of 2018 in comparison to the test interval. Generally, one can see that the test sample represents the product adequately, however small negative skewness in the tails is visible. 
Note that it is planned for future work to consider further training and testing periods to investigate the quality of the models.

\section{Hyperparameter Selection}
\label{apx:Hyperparam}
In this section of the appendix, the final hyperparameters of the \gls{ppo} algorithm after the optimization with the \gls{pbt} are presented. Thereafter, the hyperparameters of the wind and price forecast are displayed.
\subsection{Hyperparameter PPO}

\begin{table}[ht]
    \centering
    \begin{tabular}{l|l|r}
       Specification & Parameter & Value \\
        \hline
        \multirow{13}{*}{Hyperparameter \gls{ppo}}& Clip parameter & 0.432 \\
        &  Entropy coefficient & 0.001433 \\
        &  Gamma & 0 \\
        &  KL coefficient & 0.5 \\
        &  Learning rate & 0.0001 \\
        & \gls{sgd} per iteration & 7 \\
        & \gls{sgd} minibatch size & 422 \\
          & Train batch size & 2532 \\
          & Value function clip parameter & 10 \\
          & Value function loss coefficient & 0.984103\\
         &  Shared value function layers & Yes \\
                 \hline
        \multirow{4}{*}{Network specification} & First hidden layer units &  64 \\
         & Second hidden layer units &  64 \\
          & Third hidden layer units &  32 \\
          & Activation function & Tanh \\
    \end{tabular}
    \label{tab:hyperppo}
    \caption{Specifications and hyperparameter of \gls{ppo} algorithm. The hyperparameters 
     \textit{clip parameter, entropy coefficient,gamma, learning rate, \gls{sgd} per iteration} and \textit{value function loss coefficient} were selected by the \gls{pbt} algorithm.}
\end{table}

\subsection{Hyperparameter Price Forecast}

\begin{table}[h!]
    \centering
    \begin{tabular}{l|l|r}
      Specification &  Parameter & Value \\
        \hline
        \multirow{7}{*}{Hyperparameter} & Col-sample by tree & 0.817699 \\
        &Gamma & 0.097991 \\
        &Learning rate & 0.043568 \\
        &Max depth & 9 \\
       & N-estimators & 184 \\
        &Sub sample & 0.755471 \\
        \hline
        \multirow{1}{*}{Forecast Horizon} & next 15min  & 3 x 5min \\
        \hline
        \multirow{1}{*}{Forecast Update} & Recalculation of the forecast  & every 5min \\
    \end{tabular}
    \label{tab:hyperXGB}
    \caption{Forecast setting and 
    hyperparameters of the XGBoost for the price forecast. The hyperparameters were selected through cross-validation of the training data. }
\end{table}

\newpage
\begin{landscape}

\subsection{Hyperparameter Wind Forecast}
\begin{table}[ht]
    \centering
    \begin{tabular}{l|l|r}
       Specification & Parameter & Value \\
        \hline
        \multirow{4}{*}{Model architecture} &  Layer Size Encoder & 1 \\
        &  Layer Size Decoder & 1 \\
        &  Hidden Layer Size Encoder & 50 \\
        &  Hidden Layer Size Encoder & 50 \\
                \hline
        \multirow{5}{*}{Training Parameter} &  Learning rate & 1e-4 \\
        & alpha & 1e-5 \\
        & $l1_{ratio}$  & 0.1 \\
        &  L1 regularization & $\text{alpha} * l1_{ratio}$ \\
        &  L2 regularization & $\text{alpha} * (1- l1_{ratio}) * 0.5$ \\
        &  Early stopping (patience) & 5 \\
        &  Mixed teacher forcing (dynamic reduction) & 0.5-0.0 \\
                \hline
        \multirow{3}{*}{Training Specification} &  Training data & Prev. 2 months \\
         & Validation data & Last  12,5\%  of training data \\
         & Test data & Current month \\
                 \hline
         \multirow{4}{*}{Features} & U-Component of wind& 10m above the ground \\
         & U-Component of wind & 100m above the ground  \\
         & V-Component of wind & 10m above the ground \\
         & V-Component of wind & 100m above the ground \\
         \hline
         \multirow{1}{*}{Forecast horizon} &  4 hours & 16 x 15min \\
                \hline
        \multirow{1}{*}{Forecast update} &  Recalculation of forecast & Every 15min \\
                \hline
    \end{tabular}
    \label{tab:hyperwind}
    \caption{Forecast settings and hyperparameters of the wind forecast model. }
\end{table}
\end{landscape}
\end{document}